\documentclass[runningheads]{llncs}

\usepackage{eccv}

\usepackage{eccvabbrv}
\usepackage{graphicx}
\usepackage{booktabs}
\usepackage{multirow}
\usepackage{multicol}
\usepackage{pifont}
\usepackage[table]{xcolor}
\definecolor{bestblue}{HTML}{E8F0FE}    
\definecolor{oursrow}{HTML}{F3F3F3}     
\definecolor{sechead}{HTML}{EBEBEB}     
\usepackage[accsupp]{axessibility}

\usepackage{hyperref}
\usepackage{orcidlink}

\begin{document}

\title{SignBind-LLM: Multi-Stage Modality Fusion \\ for Sign Language Translation}

\titlerunning{SignBind-LLM}

\author{Marshall Thomas\orcidlink{0009-0002-5359-4060} \and
Edward Fish\orcidlink{0009-0004-2964-8430} \and
Richard Bowden\orcidlink{0000-0003-3285-8020}}

\authorrunning{M.~Thomas et al.}

\institute{University of Surrey, Guildford, GU2 7XH, United Kingdom
\email{marshall.thomas@surrey.ac.uk}\\
\url{https://www.surrey.ac.uk/centre-vision-speech-signal-processing}}

\maketitle

\begin{figure}
    \centering
    \vspace{-25pt}\includegraphics[width=1\linewidth]{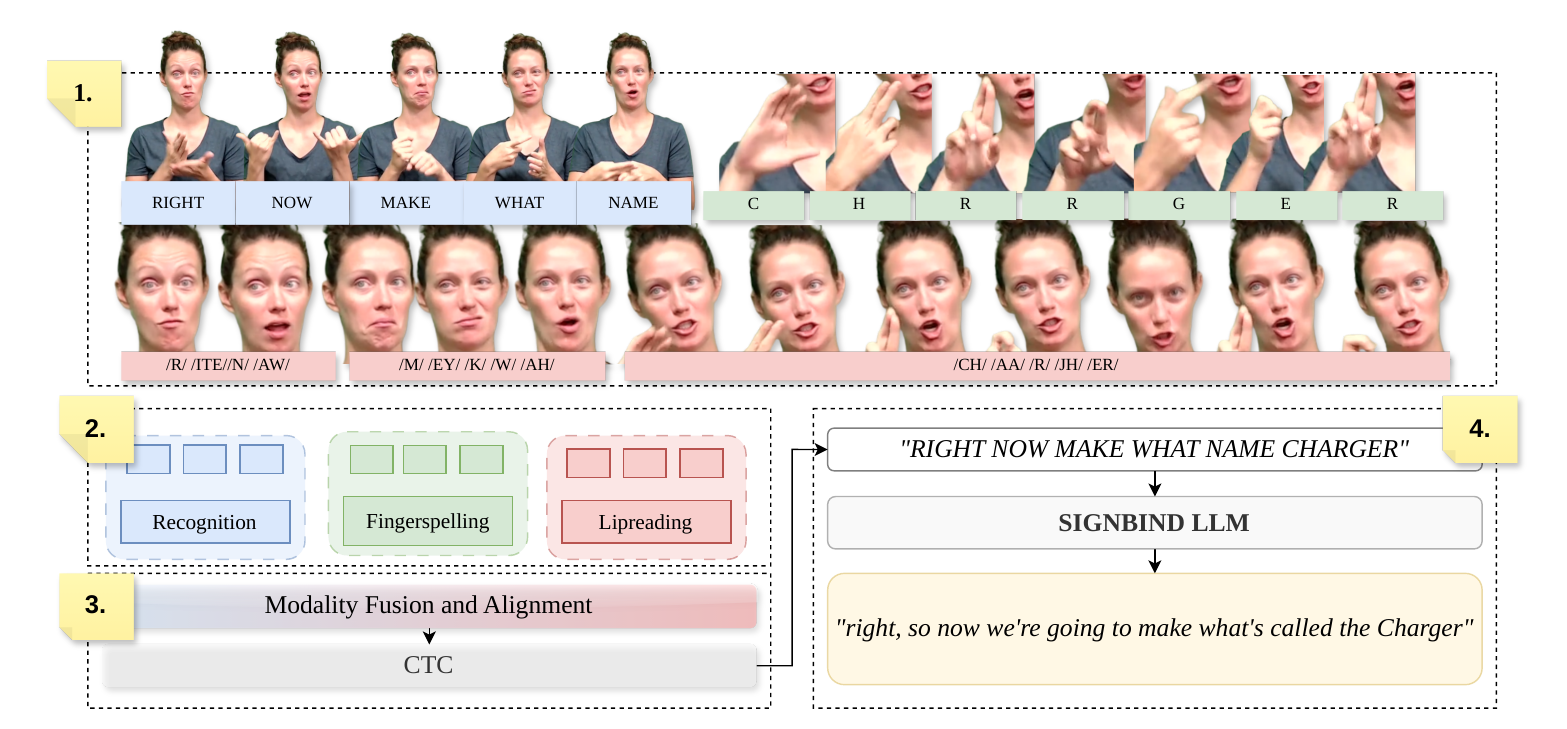}
    \caption{SignBind-LLM decomposes sign language translation into specialized sub-tasks. We train dedicated experts for continuous sign recognition, fingerspelling detection, and lipreading via CTC on automatically generated pseudo-glosses, fuse them with learned temporal alignment, and decode with a pre-trained LLM. This modular approach substantially outperforms end-to-end alternatives while requiring orders of magnitude less compute.}
    \label{fig:teaser}
\end{figure}

\vspace{-25pt}
\begin{abstract}
Current sign language translation (SLT) systems attempt to learn all aspects of signing---manual gestures, high-speed fingerspelling, and asynchronous non-manual facial cues---within a single end-to-end network. Learning multiple tasks without detailed supervision leads to poor recognition of fingerspelled proper nouns and technical terms, and leaves rich disambiguating information from lip movements largely unexploited. We introduce \textbf{SignBind-LLM}, a modular framework that addresses these limitations through three dedicated expert streams: one for continuous signing, one for fingerspelling, and one for lipreading. Each expert is pre-trained independently using CTC on approximately two million automatically generated pseudo-gloss sequences, removing the need for manual gloss annotation. A lightweight transformer with learned temporal alignment fuses the expert outputs, and a pre-trained language model translates the resulting pseudo-gloss sequences into fluent spoken English. At matched decoder scale (250M parameters), our architecture already surpasses all prior methods, confirming that the gains are architectural rather than a consequence of scaling the language model. Scaling to a larger decoder sets a new state-of-the-art across How2Sign: 23.1, BOBSL: 7.0, and ChicagoFSWild+: 73.6\%, while requiring significantly lower training cost than prior approaches.
\end{abstract}


\section{Introduction}\label{sec:intro}

Sign languages are fully-fledged natural languages used by millions of Deaf people worldwide. They rely simultaneously on manual gestures---handshapes, movements, and spatial locations---and non-manual signals such as facial expressions, mouthings, and mouth gestures\footnote{In sign language linguistics, \emph{mouthing} denotes silent articulation of spoken words accompanying signs, while \emph{mouth gestures} are non-verbal movements integral to the sign itself, conveying grammatical or affective nuance.} Crucially, these channels operate asynchronously: a signer's lip movements do not temporally coincide with their manual articulations, and fingerspelling proceeds at a fundamentally different rate than continuous signing. The field has progressed from isolated sign recognition through continuous recognition to gloss-free sign language translation (SLT), driven by large video--text datasets~\cite{YT-ASL, How2Sign, MDGS} and the integration of large language models~\cite{lin-etal-2023-gloss}. While gloss-free SLT with LLM's have shown improvement over gloss-based CTC approaches, they exhibit two major limitations that limit their real world application.

\noindent\textbf{Fingerspelling.} Fingerspelling constitutes 12--35\% of ASL~\cite{ASL_Fingerspelling} and is the primary mechanism for proper nouns, technical terms, and loanwords---precisely the content that often matters most in a translation. Current gloss-free SLT systems either treat fingerspelling as a subordinate task within the main translation pipeline~\cite{Llava-SLT, low2025sagesegmentawareglossfreeencoding} or ignore it entirely~\cite{UniSign, fish2025geosignhyperboliccontrastiveregularisation}. The consequence is systematic mistranslation of names, places, and technical vocabulary.

\noindent\textbf{Non-manual cues.}
Mouthings carry crucial disambiguating information---akin to visemes in speech~\cite{Phonemes, Visemes}---and can account for up to 40\% of the linguistic information in natural sign production~\cite{Lipreading_Sign}. Yet existing SLT models rarely utilise these signals. The central difficulty is not that the information is absent from the video, but that it is \emph{temporally misaligned} with the manual signs: mouthings may precede or lag behind the corresponding hand movements, and this offset varies continuously. Na\"ive feature concatenation at the frame level therefore fails to capture the correspondence.

\noindent
These two challenges share a common cause. In contrast to automatic speech recognition~\cite{VITA} and visual speech recognition~\cite{VALLR}, where massive pre-training on dedicated modalities is standard practice, current SLT systems force a single network to simultaneously learn continuous signing, high-speed fingerspelling, and asynchronous non-manual cues. This requires a single model to simultaneously learn three distinct recognition problems. Furthermore, training large LLMs directly on visual features introduces significant computational overhead and complicates scaling to larger models. 

To overcome these limitations, we introduce \textbf{SignBind-LLM}, a framework built on a simple principle: \emph{decompose, specialise, then fuse}. We make the following contributions:

\begin{enumerate}
    \item \textbf{Multi-stream expert architecture.} We introduce three dedicated processing streams---for continuous signing, fingerspelling, and lipreading---each pre-trained independently via CTC loss to generate independent text . A dynamic router with straight-through Gumbel-Softmax assigns video segments to the appropriate manual expert during both training and inference.
    \item \textbf{Automatic pseudo-gloss generation.} We generate and release approximately two million synthetic pseudo-gloss sequences from subtitle text using GPT-4o~\cite{GPT}, providing weak CTC supervision without any manual gloss annotation.
    \item \textbf{Temporal-aware fusion.} A cross-attention mechanism with learned projection matrices dynamically fuses the manual and non-manual streams, handling variable temporal asynchrony without fixed alignment assumptions. We show empirically that this learned alignment substantially outperforms any fixed temporal offset.
    \item \textbf{LLM decoding.} A pre-trained encoder--decoder LM translates pseudo-gloss sequences into fluent English, trained independently on text pairs and therefore fully decoupled from the visual pipeline.
\end{enumerate}

\noindent We evaluate on How2Sign~\cite{How2Sign}, ChicagoFSWild+~\cite{ChicagoFS}, and BOBSL~\cite{albanie2021bbcoxfordbritishsignlanguage}, requiring only 259 GPU-hours of training---$124\times$ less than SSVP-SLT~\cite{rust2024ssvpslt}. Critically, at matched decoder scale (250M parameters), SignBind-LLM already outperforms all prior methods (16.5 vs.\ 15.5 B-4 on How2Sign), demonstrating the gains are architectural. Scaling to a larger decoder yields further improvements ($+$7.6 B-4 on How2Sign, $+$4.4 B-4 on BOBSL, $+$2.1\,pp letter accuracy on ChicagoFSWild+), though this comparison is not parameter-matched with prior methods.

\section{Related Work}\label{sec:related}

\subsection{Sign Language Translation}

Early SLT methods relied on manually annotated glosses as an intermediate representation. Gloss-based pipelines~\cite{camgoz2020sign, zhao2023conditionalvariationalautoencodersign, zhang2023sltunetsimpleunifiedmodel} use CTC-based continuous sign language recognition (CSLR)~\cite{CTC, min2021visual, koller2020weakly} to predict gloss sequences, which are then translated to spoken language via sequence-to-sequence models~\cite{Glosses}. While effective on controlled benchmarks, these approaches inherit errors from the gloss recogniser and require costly aligned annotations that are unavailable for most sign languages.

Gloss-free methods bypass this bottleneck by learning direct video-to-text mappings. Large-scale video--text pre-training~\cite{zhou2023gfslv, lin-etal-2023-gloss, zhou2024scalingmultimodalpretrainingsign, tarrés2023signlanguagetranslationinstructional} and contrastive objectives~\cite{gueuwou2024signmusketeersefficientmultistreamapproach} have improved translation fluency considerably. Recent work has focused on leveraging LLMs for decoding sign features, whether from RGB~\cite{YT-ASL}, pose~\cite{fish2025geosignhyperboliccontrastiveregularisation}, or fused modalities~\cite{UniSign, wong2024sign2gpt}. However, all of these systems attempt to learn signing, fingerspelling, and non-manual cues within a single network, leading to systematic underperformance on fingerspelled content and underuse of facial information.

Several multimodal approaches have fused RGB, pose, and optical flow~\cite{fish2025geosignhyperboliccontrastiveregularisation, jang2025losttranslationcontextsign}, and fingerspelling-specific methods have explored synthetic augmentation~\cite{Synthetic}, large-scale in-the-wild collections~\cite{Zhidebayeva2024, georg2024fsboard3millioncharacters}, and pose-based models~\cite{fayyazsanavi2024posenet}. None of these integrates lipreading, despite the well-documented co-articulation between handshapes and lip movements in natural signing~\cite{Lipreading_Sign} and the persistent errors that arise from handshape ambiguity alone.

\subsection{Visual Speech Recognition for Signing}

Visual speech recognition (VSR) has matured rapidly through end-to-end architectures such as LipNet~\cite{assael2016lipnet}, transformer-based models like Lipformer~\cite{LipFormer}, and large-scale audiovisual pre-training with AV-HuBERT~\cite{shi2022avhubert, VALLR}. More recent work has integrated LLM-based encoders to improve contextual transcription~\cite{yeo2024visualspeechmeetslanguage}, achieving strong performance even under challenging conditions. In the context of sign language, LLMs have also been used for fingerspelling detection~\cite{tanzer2024fingerspelling, shi2023americansignlanguageprocessing} by aligning hand gestures with textual representations.

Despite the maturity of VSR and the availability of large sign language datasets, no prior work has combined lipreading with both continuous sign recognition and fingerspelling detection in a unified framework for SLT---let alone one that explicitly accounts for the temporal misalignment between these streams. Our work fills this gap.

\section{Method}\label{sec:method}

\begin{figure*}[t]
\centering
\includegraphics[width=\textwidth]{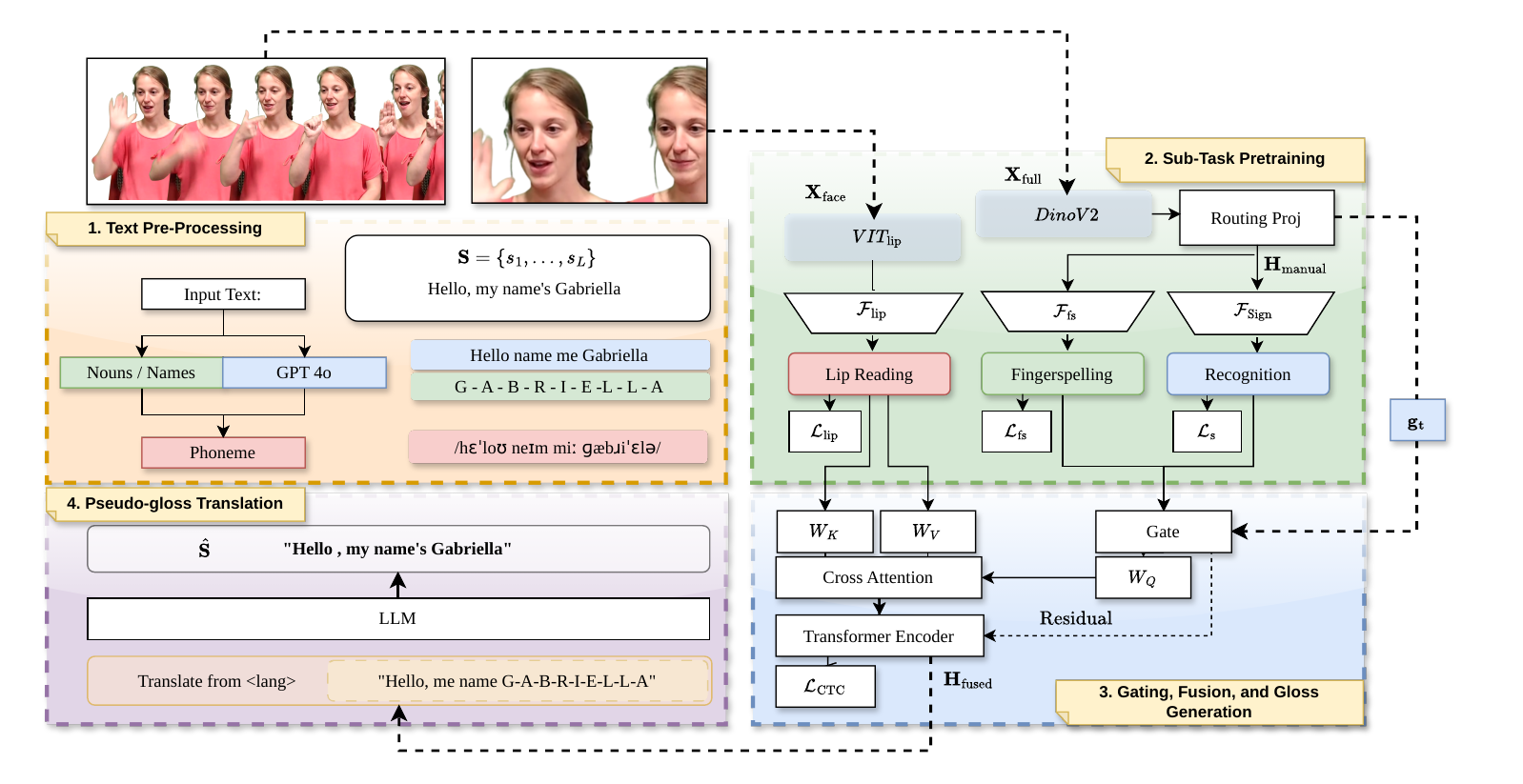}
\caption{Overview of SignBind-LLM. Stage~1: Pseudo-glosses and phonemes are generated automatically from subtitles; full-frame and face-cropped video streams are extracted in parallel. Stage~2: specialized expert networks for signing, fingerspelling, and lipreading are pre-trained independently via CTC; a dynamic segment classifier routes video frames to the appropriate manual expert. Stage~3: the expert outputs are gated by the router and fused via cross-attention with learned temporal alignment. Stage~4: a fine-tuned encoder--decoder LLM translates the fused pseudo-gloss sequence into spoken English.}
\label{fig:overview}
\end{figure*}

We propose \textbf{SignBind-LLM}, a sign language translation framework that decomposes the complex translation task into specialized sub-tasks before fusion. The key observation is that sign languages employ distinct communication channels---continuous signing, fingerspelling, and mouthings---that operate at different temporal scales and are asynchronous with respect to one another. Attempting to learn all of these within a single encoder conflates fundamentally different recognition problems. Instead, we assign each channel to a dedicated expert, fuse their outputs with learned temporal alignment, and decode with a pre-trained language model. The pipeline consists of four stages: (1)~target generation, (2)~modality-specific pre-training, (3)~multi-modal fusion, and (4)~language model decoding. Figure~\ref{fig:overview} illustrates the complete architecture.

\subsection{Problem Formulation}\label{sec:formulation}

Given a sign language video $\mathbf{X} = \{x_1, \ldots, x_T\}$ with $x_t \in \mathbb{R}^{H \times W \times 3}$, our objective is to produce an English translation $\mathbf{S} = \{s_1, \ldots, s_L\}$. We decompose this into four stages: target-generation, parallel expert encoding, fusion, and language model decoding:
\begin{equation}
    \mathbf{S} = \text{LLM}\!\Big(\text{Fuse}\big(\mathcal{E}_{\text{sign}}(\mathbf{X}),\; \mathcal{E}_{\text{fs}}(\mathbf{X}),\; \mathcal{E}_{\text{lip}}(\mathbf{X})\big)\Big)
\end{equation}
where $\mathcal{E}_{\text{sign}}$, $\mathcal{E}_{\text{fs}}$, and $\mathcal{E}_{\text{lip}}$ denote the expert encoders for continuous signing, fingerspelling, and lipreading respectively, Fuse reconciles their asynchronous outputs into a unified pseudo-gloss representation $\hat{\mathbf{G}}$, and the LLM translates $\hat{\mathbf{G}}$ into the target sentence.

\subsection{Stage 1: Target Generation}\label{sec:targets}

A significant barrier to multi-stream modelling is the lack of parallel, granular annotations. We overcome this by automatically generating three complementary training targets from existing English subtitles $\mathbf{S} = \{s_1, \ldots, s_L\}$.

\noindent\textbf{Pseudo-gloss generation.} We employ GPT-4o~\cite{GPT} to transform English sentences into sign-language-ordered pseudo-glosses:
\begin{equation}
    \mathbf{G} = \text{LLM}_{\text{gloss}}(\mathbf{S}) = \{G_1, \ldots, G_M\}
\end{equation}
where $M \leq L$ reflects the removal of semantically redundant elements such as articles and the reordering toward subject--object--verb structure. We generate approximately two million such sequences across all training corpora.

\noindent\textbf{Phoneme extraction.} To supervise the lipreading module, we extract phoneme sequences from the pseudo-glosses $\mathbf{G}$  using an off-the-shelf phonemiser: $\mathbf{P} = \text{Phonemize}(\mathbf{G}) = \{p_1, \ldots, p_N\}$, where each $p_i$ represents an English phoneme.

\noindent\textbf{Fingerspelling labels.} We use the ChicagoFSWild+~\cite{ChicagoFS} training set, decomposing word labels into character sequences $\mathbf{F} = \{f_1, \ldots, f_n\}$ where each $f_i$ is a letter of the alphabet. We additionally identify potential fingerspelling segments in the pseudo-glosses via rule-based detection of proper nouns and technical terms, which are typically fingerspelled in both ASL and BSL.

\subsection{Stage 2: Modality-Specific Encoders}\label{sec:encoders}

This stage defines the expert networks that learn to recognise each communication channel independently.

\noindent\textbf{Video pre-processing.} We process the input video into two streams using MediaPipe~\cite{MediaPipe}: full frames $\mathbf{X}_{\text{full}}$ for manual sign recognition and tightly-cropped face regions $\mathbf{X}_{\text{face}}$ for lipreading. Both streams are normalised to $224 \times 224$ resolution.

\noindent\textbf{Shared backbone and manual experts.} The core of our visual encoder is a DINOv2~\cite{oquab2024dinov2learningrobustvisual} backbone that processes $\mathbf{X}_{\text{full}}$ in 16-frame segments, producing shared features $\mathbf{H}_{\text{manual}} \in \mathbb{R}^{B \times T \times D}$. These features serve as input to three parallel components: a dynamic segment router, a continuous sign recognition head, and a fingerspelling detection head.

\noindent\textbf{Dynamic segment routing.} Applying all expert heads to all frames is both inefficient and a source of noise---for instance, the fingerspelling head will produce spurious outputs on frames containing continuous signing. We therefore learn a lightweight MLP classifier that routes each temporal segment to one of $\{\text{Sign}, \text{Fingerspelling}, \text{Rest}\}$ and labelled as `routing proj' in Fig.~\ref{fig:overview}. We first compute a segment-level representation $\bar{\mathbf{h}}_t \in \mathbb{R}^{D}$ by averaging the DINOv2 features over the frames in segment $t$, then compute class logits:
\begin{equation}
    \bar{\mathbf{h}}_t = \frac{1}{|\mathcal{S}_t|}\sum_{i \in \mathcal{S}_t} \mathbf{H}_{\text{manual},i}\,,\qquad
    \hat{\mathbf{c}}_t = \mathbf{W}_{\text{cls}}\, \bar{\mathbf{h}}_t + \mathbf{b}_{\text{cls}} \;\in\; \mathbb{R}^{3}
\end{equation}
where $\mathcal{S}_t$ denotes the set of frame indices belonging to segment $t$.

To obtain discrete routing decisions while maintaining gradient flow during training, we use the straight-through Gumbel-Softmax estimator~\cite{jang2017categoricalreparameterizationgumbelsoftmax}. The classifier achieves 85\% accuracy on this three-way task. Segment labels are assigned as follows: \textit{Fingerspelling} segments are sourced from ChicagoFSWild+ annotations; \textit{Sign} segments are those falling within a non-empty pseudo-gloss region; and \textit{Rest} segments correspond to background or transition frames. In the forward pass, a hard one-hot vector is produced via $\text{argmax}$; in the backward pass, gradients are computed through the continuous softmax relaxation:
\begin{equation}
    \mathbf{r}_t = \text{ST-GumbelSoftmax}(\hat{\mathbf{c}}_t,\, \tau) \;\in\; \{0,1\}^{3}
\end{equation}
where $\mathbf{r}_t = [r_{t}^{\text{sign}},\; r_{t}^{\text{fs}},\; r_{t}^{\text{rest}}]$ and $\tau$ is annealed from 1.0 to 0.1 over the course of training. At inference, $\text{argmax}$ is applied directly. The routing vector $\mathbf{r}_t$ is passed to the fusion stage (Stage~3) to gate the expert outputs.

\noindent\textbf{Continuous sign recognition.} In parallel with the router, the continuous signing expert processes the DINOv2 features. We add learned positional embeddings $\mathbf{PE} \in \mathbb{R}^{T \times D}$ to obtain expert features $\mathbf{F}_{\text{sign}} \in \mathbb{R}^{B \times T \times D}$, then project through a CTC head to the sign vocabulary $\mathcal{V}_{\text{sign}}$:
\begin{equation}
    \mathbf{F}_{\text{sign}} = \mathbf{H}_{\text{manual}} + \mathbf{PE}\,,\qquad
    \hat{\mathbf{L}}_{\text{sign}} = \mathbf{F}_{\text{sign}}\,\mathbf{W}_{\text{sign}} + \mathbf{b}_{\text{sign}} \;\in\; \mathbb{R}^{B \times T \times |\mathcal{V}_{\text{sign}}|}
\end{equation}

\noindent\textbf{Fingerspelling detection.} The fingerspelling expert shares the same position-encoded features but uses an independent CTC head projecting to the 26 characters of the English alphabet:
\begin{equation}
    \mathbf{F}_{\text{fs}} = \mathbf{H}_{\text{manual}} + \mathbf{PE}\,,\qquad
    \hat{\mathbf{L}}_{\text{fs}} = \mathbf{F}_{\text{fs}}\,\mathbf{W}_{\text{fs}} + \mathbf{b}_{\text{fs}} \;\in\; \mathbb{R}^{B \times T \times 26}
\end{equation}

\noindent\textbf{Lipreading module.} The lipreading expert processes the face-cropped stream $\mathbf{X}_{\text{face}}$ independently. Note that this stream captures \emph{mouthings}---speech-derived lip movements that co-occur with signs---rather than the full range of sign-language mouth gestures (which include non-speech movements integral to grammatical marking). This is a deliberate design choice: mouthings carry phonemic information that can be directly supervised with phoneme targets, whereas mouth gestures would require separate annotation. We employ a masked ViT ($\text{ViT}_{\text{lip}}$) with a 50\% masking ratio to force robust representation learning as shown to be effective in~\cite{VALLR}. A 1D convolution then performs temporal downsampling to produce expert features $\mathbf{F}_{\text{lip}}$, and a final linear CTC head projects to the phoneme vocabulary $\mathcal{P}$:
\begin{equation}
\begin{aligned}
    \mathbf{H}_{\text{lip}} &= \text{ViT}_{\text{lip}}(\text{Mask}(\mathbf{X}_{\text{face}},\, 0.5)) \;\in\; \mathbb{R}^{B \times T \times D_{\text{lip}}}\\
    \mathbf{F}_{\text{lip}} &= \text{Conv1D}(\mathbf{H}_{\text{lip}}) \;\in\; \mathbb{R}^{B \times T' \times D'}\\
    \hat{\mathbf{L}}_{\text{lip}} &= \mathbf{F}_{\text{lip}}\,\mathbf{W}_{\text{lip}} + \mathbf{b}_{\text{lip}} \;\in\; \mathbb{R}^{B \times T' \times |\mathcal{P}|}
\end{aligned}
\end{equation}
where $T' < T$ due to the convolutional stride, $D_{\text{lip}}$ is the ViT hidden dimension, and $D'$ is the dimension of the temporally downsampled features.

All three CTC heads ($\hat{\mathbf{L}}_{\text{sign}}, \hat{\mathbf{L}}_{\text{fs}}, \hat{\mathbf{L}}_{\text{lip}}$) and the segment classifier are trained independently using CTC loss on the targets generated in Stage~1. During fusion (Stage~3), the CTC heads are discarded and the penultimate expert features ($\mathbf{F}_{\text{sign}}, \mathbf{F}_{\text{fs}}, \mathbf{F}_{\text{lip}}$) are used directly.

\subsection{Stage 3: Temporal-Aware Multi-Modal Fusion}\label{sec:fusion}

This stage reconciles the asynchronous expert outputs into a unified representation. The CTC heads used during pre-training are discarded, and the penultimate expert embeddings---which retain richer representational structure than the low-dimensional logits---are projected to a common dimension $d$ via learned linear maps $\phi_{\text{sign}}: \mathbb{R}^{D} \rightarrow \mathbb{R}^d$, $\phi_{\text{fs}}: \mathbb{R}^{D} \rightarrow \mathbb{R}^d$, $\phi_{\text{lip}}: \mathbb{R}^{D'} \rightarrow \mathbb{R}^d$. The lipreading features are temporally upsampled from $T'$ to $T$ via linear interpolation:
\begin{equation}
    \mathbf{E}_{\text{sign}} = \phi_{\text{sign}}(\mathbf{F}_{\text{sign}}),\;\;
    \mathbf{E}_{\text{fs}} = \phi_{\text{fs}}(\mathbf{F}_{\text{fs}}),\;\;
    \mathbf{E}_{\text{lip}} = \phi_{\text{lip}}(\text{Upsample}(\mathbf{F}_{\text{lip}},\, T))
\end{equation}
where $\mathbf{E}_{\text{sign}}, \mathbf{E}_{\text{fs}}, \mathbf{E}_{\text{lip}} \in \mathbb{R}^{B \times T \times d}$.

\noindent\textbf{Gated manual aggregation.} The routing vector $\mathbf{r}_t$ from Stage~2 selects between expert outputs at each time step, producing a unified manual representation:
\begin{equation}
    \mathbf{M}_t = r_{t}^{\text{sign}} \cdot \mathbf{E}_{\text{sign},t} \;+\; r_{t}^{\text{fs}} \cdot \mathbf{E}_{\text{fs},t} \;+\; r_{t}^{\text{rest}} \cdot \mathbf{e}_{\text{null}}
\end{equation}
where $\mathbf{e}_{\text{null}} \in \mathbb{R}^d$ is a learned null embedding representing rest periods. Since $\mathbf{r}_t$ is one-hot in the forward pass, exactly one branch is active at each time step; gradients still flow to all branches through the Gumbel-Softmax relaxation during this training stage.

\noindent\textbf{Cross-attention fusion.} The relative importance of manual signs versus mouthings is strongly context-dependent. During continuous signing, the manual stream carries the primary semantic load; during fingerspelling, where individual handshapes are visually ambiguous, mouthings become the dominant disambiguating signal. We propose a two-level fusion design: first, a \emph{gated manual aggregation} (described above) that routes between manual experts---itself a contribution that substantially outperforms naïve concatenation (see Table~\ref{tab:fusion_pretrain})---and second, a \emph{cross-attention} mechanism that integrates the lipreading stream. We model this dynamic balance via multi-head cross-attention, where the choice of query and key/value sources encodes a specific inductive bias: each manual-stream position \emph{queries} the lipreading stream to retrieve the most relevant non-manual context, rather than the reverse. This reflects the linguistic structure of signing, where manual articulations form the primary lexical channel and mouthings provide supporting disambiguation~\cite{DGS_Fingerspelling}.

Concretely, we compute $N_h$ parallel attention heads. For each head $i$, we project the gated manual representation and the lipreading features into queries, keys, and values of dimension $d_h = d / N_h$:
\begin{equation}
    \mathbf{Q}^{(i)} = \mathbf{M}\mathbf{W}_Q^{(i)}, \quad \mathbf{K}^{(i)} = \mathbf{E}_{\text{lip}}\mathbf{W}_K^{(i)}, \quad \mathbf{V}^{(i)} = \mathbf{E}_{\text{lip}}\mathbf{W}_V^{(i)}
\end{equation}
where $\mathbf{W}_Q^{(i)}, \mathbf{W}_K^{(i)}, \mathbf{W}_V^{(i)} \in \mathbb{R}^{d \times d_h}$. The attention weight matrix $\mathbf{A}^{(i)} \in \mathbb{R}^{T \times T}$ is then:
\begin{equation}
    \mathbf{A}^{(i)} = \text{Softmax}\!\left(\frac{\mathbf{Q}^{(i)} {\mathbf{K}^{(i)}}^\top}{\sqrt{d_h}}\right)
\end{equation}
Entry $A^{(i)}_{t,t'}$ measures how strongly manual-stream position $t$ attends to lipreading position $t'$. Crucially, no constraint is placed on the temporal relationship between $t$ and $t'$: the model is free to learn that, for instance, the mouthing corresponding to a sign at frame $t$ actually occurs several frames earlier at $t' < t$. This is the mechanism by which the architecture handles the variable asynchrony between channels without requiring a fixed temporal offset.

The $N_h$ heads are concatenated and projected back to dimension $d$ via $\mathbf{W}_O \in \mathbb{R}^{d \times d}$, then combined with a residual connection, layer normalisation (LN), and a position-wise feed-forward network:
\begin{equation}
\begin{aligned}
    \mathbf{H}' &= \mathbf{M} + \mathbf{W}_O\,[\text{head}_1;\,\ldots\,;\text{head}_{N_h}] \\
    \mathbf{H}_{\text{fused}} &= \text{LN}(\mathbf{H}') + \text{FFN}(\text{LN}(\mathbf{H}'))
\end{aligned}
\end{equation}
where $\text{head}_i = \mathbf{A}^{(i)}\mathbf{V}^{(i)}$. Different heads can specialise to different temporal offsets or linguistic functions---for example, one head may learn to align mouthings with the onset of the corresponding sign, while another captures co-articulatory overlap between consecutive signs.

\noindent\textbf{Temporal modelling.} The fused features $\mathbf{H}_{\text{fused}}$ are passed through a shallow transformer encoder to model long-range temporal dependencies and to resolve any residual misalignment between the expert streams:
\begin{equation}
    \mathbf{Z} = \text{TransformerEncoder}(\mathbf{H}_{\text{fused}}) \;\in\; \mathbb{R}^{B \times T \times d}
\end{equation}

\noindent The entire fusion module is trained via CTC loss to produce the correct pseudo-gloss sequence:
\begin{equation}
    \hat{\mathbf{G}} = \text{CTC-Decode}(\text{Softmax}(\mathbf{W}_{\text{out}}\mathbf{Z} + \mathbf{b}_{\text{out}}))
\end{equation}

\subsection{Stage 4: Language Model Decoding}\label{sec:lm}

The fused representation $\hat{\mathbf{G}}$ is a sequence of pseudo-glosses that follows sign-language word order and lacks function words. It is not a well-formed English sentence. The final stage employs a fine-tuned LLM (Flan-T5-XL~\cite{Flan-T5}) to translate this intermediate representation into a spoken English caption: $\hat{\mathbf{S}} = \text{LLM}_\theta(\hat{\mathbf{G}})$.

This model is trained \emph{independently} of the visual pipeline on a large corpus of (pseudo-gloss, English sentence) pairs $(\mathbf{G}, \mathbf{S})$ generated in Stage~1, using standard cross-entropy loss. The decoupled design has an important practical benefit: because the LLM never sees visual features during training, it learns a robust linguistic mapping that generalises to the noisy, imperfect pseudo-gloss outputs produced by the visual model at inference time. This decoupling also simplifies training and enables flexible deployment with different LLM sizes. It also enables researchers to simply update any component of the architecture without needing to retrain the LLM which has significant benefits in production. 

\subsection{Training Strategy}\label{sec:training}

Training proceeds in strictly staged phases to ensure stable convergence and component specialisation. 

\textbf{Phase~1 (Pre-training):} (i)~Each expert head and the segment classifier are trained independently on large-scale data (YouTube-ASL~\cite{YT-ASL} for sign recognition, ChicagoFSWild+~\cite{ChicagoFS} for fingerspelling, and BOBSL~\cite{albanie2021bbcoxfordbritishsignlanguage} for BSL) using their respective CTC losses. (ii)~Expert weights are then \textit{frozen} and the fusion module is trained on their outputs, again via CTC. (iii)~Separately, the LLM is pre-trained on all available $(\mathbf{G}, \mathbf{S})$ text pairs. 

\textbf{Phase~2 (Fine-tuning):} The same staged sequence is repeated on the target evaluation dataset (\emph{e.g.}\ How2Sign): experts are fine-tuned first, then frozen, then the fusion module is fine-tuned on their updated outputs. The LLM remains frozen throughout Phase~2, having already learned a robust pseudo-gloss-to-English mapping. This decoupled curriculum ensures each component becomes a strong specialist at its sub-task before its outputs are consumed by downstream modules, and avoids the expensive joint optimisation over all parameters that end-to-end methods require.

\section{Experiments}\label{sec:experiments}

\subsection{Datasets}\label{sec:datasets}

We evaluate on two sign languages (ASL and BSL) across three benchmarks.

\noindent\textbf{YouTube-ASL~\cite{YT-ASL}.} A large-scale ASL corpus of 60K videos (${\sim}$1,000\,h) from over 2,500 signers across diverse backgrounds, lighting conditions, and signing styles. The content is not interpreter-mediated, preserving the diversity of natural signing. We use YouTube-ASL for visual--language pre-training.

\noindent\textbf{How2Sign~\cite{How2Sign}.} A continuous ASL dataset derived from instructional videos, totalling over 80\,h across green-screen and studio settings. Eleven signers produced 35,191 sentence-level clips (avg.\ 162 frames, 17 words) with a vocabulary exceeding 16,000 English words. We fine-tune and evaluate on the standard How2Sign train/test splits.

\noindent\textbf{ChicagoFSWild+~\cite{ChicagoFS}.} An in-the-wild fingerspelling dataset containing 55,232 sequences from 260 signers, with varied environments and frequent occlusions. We use the standard train/test split for both pre-training the fingerspelling head and evaluation.

\noindent\textbf{BOBSL~\cite{albanie2021bbcoxfordbritishsignlanguage}.} A BSL dataset sourced from BBC broadcast footage, comprising 1,467\,h across 1,962 episodes interpreted by 39 signers, with ${\sim}$1.2M weakly-aligned English subtitle sentences. We use the improved alignments generated from the method described in ~\cite{alignments} and the train/test splits as defined in ~\cite{albanie2021bbcoxfordbritishsignlanguage}.

\subsection{Metrics}

We report \textbf{BLEU-1} (B-1) and \textbf{BLEU-4} (B-4)~\cite{BLEU} for translation quality, \textbf{ROUGE-L} (R-L)~\cite{ganesan2018rouge20updatedimproved} for recall-oriented overlap, and \textbf{Letter Accuracy} (L-Acc.)~\cite{WER} for fingerspelling recognition. Higher is better throughout.

\subsection{Implementation Details}

All  experiments use Flan-T5-XL (3.0B) as the LLM decoder unless otherwise stated. The DINOv2-Base backbone processes 16-frame segments at $224 \times 224$ resolution. The fusion transformer has 4 layers with 8 attention heads and dimension $d{=}512$. The Gumbel-Softmax temperature $\tau$ is annealed from 1.0 to 0.1 over 50K steps. We use AdamW with cosine learning rate scheduling. Total training cost is 259 GPU-hours on a single A100 GPU. Code, all generated pseudo-gloss data, and all pre-trained models will be released on publication.

\section{Results}\label{sec:results}

\begin{table*}[!t]
\centering
\caption{\textbf{Sign language translation on How2Sign.}
SignBind-LLM\textsubscript{base} (250M) already outperforms all prior methods---including those with 582M decoders and additional pose input. When using the same decoder size as prior work (Flan-T5-Base, 250M), our method already improves BLEU-4 from 15.5 to 16.5---demonstrating that the multi-stream architecture, not merely a larger LM, drives the improvement.
Scaling to Flan-T5-XL (3.0B) sets a new SOTA with $+$7.6 B-4 over SSVP-SLT.
\underline{Underline}: previous best.\;\colorbox{bestblue}{\strut\textbf{Blue}}: best overall.}
\label{tab:how2sign}
\vspace{-4pt}
\scriptsize
\setlength{\tabcolsep}{4pt}
\renewcommand{\arraystretch}{1.06}
\begin{tabular}{@{}l l r cc ccc r@{}}
\toprule
\multirow{2}{*}{\textbf{Method}} &
  \multirow{2}{*}{\textbf{LM Decoder}} &
  \multirow{2}{*}{\textbf{\#P}} &
  \multicolumn{2}{c}{\textbf{Input}} &
  \multicolumn{3}{c}{\textbf{How2Sign Test}} &
  \multirow{2}{*}{\textbf{Hrs}} \\
\cmidrule(lr){4-5}\cmidrule(lr){6-8}
 & & & \textbf{P} & \textbf{R}
 & \textbf{B-1}
 & \textbf{B-4}
 & \textbf{R-L} & \\
\midrule
GloFE-VN~\cite{lin-etal-2023-gloss}
  & --       & --   & \checkmark &            & 14.9 &  2.2 & 12.6 & -- \\
MSLU~\cite{zhou2024scalingmultimodalpretrainingsign}
  & mT5-B & 582M & \checkmark & \checkmark & 20.1 &  2.4 & 17.2 & -- \\
SLT-IV~\cite{tarrés2023signlanguagetranslationinstructional}
  & Trans. & ${\approx}$30M & & \checkmark & 34.0 &  8.0 &  -- & -- \\
C\textsuperscript{2}RL~\cite{chen2024c2rlcontentcontextrepresentation}
  & mBART    & 680M &            & \checkmark & 29.1 &  9.4 & 27.0 & -- \\
FLa-LLM~\cite{chen2024factorizedlearningassistedlarge}
  & mBART    & 680M &            & \checkmark & 29.8 &  9.7 &  -- & -- \\
\addlinespace[2pt]
YouTube-ASL~\cite{YT-ASL}
  & T5-B  & 250M & \checkmark &            & 37.8 & 12.4 &  -- & -- \\
SignMusk.~\cite{gueuwou2024signmusketeersefficientmultistreamapproach}
  & T5-B  & 250M &            & \checkmark & 41.5 & 14.3 &  -- & 880 \\
Uni-Sign~\cite{UniSign}
  & mT5-B & 582M & \checkmark & \checkmark & 40.2 & 14.9 & 36.0 & -- \\
Geo-Sign~\cite{fish2025geosignhyperboliccontrastiveregularisation}
  & mT5-B & 582M & \checkmark &            & 40.8 & 15.1 & 35.4 & -- \\
SSVP-SLT~\cite{rust2024ssvpslt}
  & T5-B  & 250M &            & \checkmark
  & \underline{43.2} & \underline{15.5} & \underline{38.4} & 32k \\
\midrule
\rowcolor{oursrow}
Ours\textsubscript{base}
  & FT5-B  & 250M &  & \checkmark & 45.0 & 16.5 & 31.5 & \multirow{3}{*}{\textbf{216/?}} \\
\rowcolor{oursrow}
Ours\textsubscript{large}
  & FT5-L & 780M &  & \checkmark & 46.7 & 18.6 & 36.9 & \\
\rowcolor{bestblue}
\textbf{Ours\textsubscript{xl}}
  & \textbf{FT5-XL} & \textbf{3.0B} & & \checkmark
  & \textbf{50.2} & \textbf{23.1} & \textbf{42.7} & 216/43 \\
\bottomrule
\end{tabular}\\[2pt]
{\scriptsize \#P = LM decoder parameters.\; P = Pose, R = RGB.\; FT5 = Flan-T5.\; B-1/B-4 = BLEU-1/4.\; R-L = ROUGE-L.\; Hrs = Visual 
 Encoder GPU hours/LLM GPU-hours.}
\end{table*}

\subsection{Sign Language Translation}

\noindent\textbf{How2Sign.}
Table~\ref{tab:how2sign} compares SignBind-LLM against recent methods. At the \emph{base} scale (Flan-T5-Base, 250M), our model already achieves 16.5 B-4, surpassing the previous SOTA (SSVP-SLT, 15.5 B-4) with an identically-sized decoder---demonstrating that the gains arise from the multi-stream architecture, not a larger LM. However, we also show that our model scales gracefully as we increase the LLM size. For example, Flan-T5-XL (3.0B) yields \textbf{50.2} B-1, \textbf{23.1} B-4, and \textbf{42.7} R-L ($+$7.6 B-4 over SSVP-SLT), with significantly less training cost. Our method uses 216 A100 GPU hours to train the visual encoders and just 43 hours of fine-tuning on the LLM. The advantage of our decomposed approach is that the visual encoders can be trained just once and then are interchangeable with other LLM sizes (as shown in the table and supplementary).

\noindent\textbf{BOBSL.}
Table~\ref{tab:bobsl} presents results on BOBSL. SignBind-LLM achieves \textbf{7.0} B-4 and \textbf{25.8} R-L, improving B-4 by $+$4.4 over the previous best~\cite{jang2025losttranslationcontextsign}. These gains confirm that multi-stream fusion generalises from ASL to BSL without architectural modification.

\begin{table*}[!t]
\centering
\scriptsize
\setlength{\tabcolsep}{4pt}
\renewcommand{\arraystretch}{1.06}
\begin{minipage}[t]{0.47\linewidth}
\centering
\captionof{table}{Translation on \textbf{BOBSL}. B-1/B-4 = BLEU-1/4, R-L = ROUGE-L.}
\label{tab:bobsl}

\begin{tabular}{@{}l ccc@{}}
\toprule
\textbf{Method} & \textbf{B-1} & \textbf{B-4} & \textbf{R-L} \\
\midrule
GFSLT~\cite{zhou2023gfslv}       & --   & 0.6 &  7.4 \\
Sign2GPT~\cite{wong2024sign2gpt} & --   & 0.9 & 11.4 \\
Albanie~et~al.~\cite{albanie2021bbcoxfordbritishsignlanguage}
                                  & 12.8 & 1.0 & 10.2 \\
Sincan~et~al.~\cite{sincan2023contextneedscalingneural}
                                  & 18.8 & 1.3 &  8.9 \\
Lost in Trans.~\cite{jang2025losttranslationcontextsign}
                                  & --   & \underline{2.6} & \underline{15.6} \\
\midrule
\rowcolor{bestblue}
\textbf{Ours}
   & \textbf{28.5} & \textbf{7.0} & \textbf{25.8} \\
\bottomrule
\end{tabular}
\end{minipage}
\hfill
\begin{minipage}[t]{0.47\linewidth}
\centering
\captionof{table}{Fingerspelling on \textbf{ChicagoFSWild+} test set. L-Acc = letter accuracy.}
\label{tab:chicago}
\begin{tabular}{@{}l c@{}}
\toprule
\textbf{Method} & \textbf{L-Acc} \\
\midrule
FS Recognition~\cite{shi2018aslfswild}                & 41.2 \\
Iterative Attn.~\cite{shi2019iterativeattention}      & 46.7 \\
TDC-SL~\cite{papadimitriou2020tdcsl}                  & 50.0 \\
CtoML~\cite{li2023ctoml}                              & 54.9 \\
FS Attention~\cite{kabade2023aslfsattention}           & 57.8 \\
FSS-Net~\cite{shi2022fssnet}                          & 64.4 \\
FS PoseNet~\cite{fayyazsanavi2024posenet}              & \underline{71.1} \\
\midrule
\rowcolor{bestblue}
\textbf{Ours}                                 & \textbf{73.6} \\
\bottomrule
\end{tabular}
\end{minipage}
\end{table*}

\subsection{Qualitative Analysis}

Table~\ref{tab:qualitative} presents representative translations. Examples~1--2 show how noisy pseudo-glosses from individual experts are resolved into coherent English after fusion incorporates lipreading phonemes and the LLM adds function words. Example~3 illustrates a failure mode: the ambiguous fused glosses \texttt{TODAY WE WORK LOWER HOUSE} cause the LLM to hallucinate ``basement'' instead of the reference ``lower body''---a limitation of the two-stage design where the LLM conditions only on intermediate text.

\subsection{Fingerspelling Recognition}
Table~\ref{tab:chicago} reports letter accuracy on ChicagoFSWild+. SignBind-LLM achieves \textbf{73.6\%}, a 2.1\,pp improvement over FS PoseNet~\cite{fayyazsanavi2024posenet}, despite not being specifically designed as a fingerspelling system. The gains arise from lipreading cues during fusion that disambiguate visually similar handshapes; removing the lipreading stream causes an 18.9\,pp drop as discussed in Sec.~\ref{sec:ablation}. Precision and recall on ChicagoFSWild+ are \textbf{76.2\%} and \textbf{70.4\%} respectively. The benefit of dedicated fingerspelling handling is further confirmed by a part-of-speech analysis (supplementary): proper noun accuracy (PROPN) reaches 0.56 for SignBind-LLM versus 0.40 for C\textsuperscript{2}RL and 0.28 for Geo-Sign---a 40\% relative improvement---directly attributable to fingerspelling recognition of names, locations, and technical terms.

\begin{table}[]
\centering
\caption{\textbf{Qualitative examples} from How2Sign (1, 3) and BOBSL (2). \textit{Fused} = pseudo-gloss output from the fusion stage $\mathbf{H}_\text{fused}$; \textit{Ours} = final LLM translation; \textit{Ref} = ground-truth subtitle. Examples~1--2 succeed; Example~3 shows semantic drift. Extended outputs in supplementary.}
\label{tab:qualitative}
\vspace{-4pt}
\footnotesize
\setlength{\tabcolsep}{3pt}
\renewcommand{\arraystretch}{1.04}
\begin{tabular}{@{} c p{0.87\columnwidth} @{}}
\toprule
\rowcolor{sechead}
\multicolumn{2}{@{}l}{\textbf{Example 1} --- How2Sign \textcolor{green!60!black}{\ding{51}}} \\
\textit{F} & \texttt{\small GAIN BALANCE ON TOE YOU HAVE CENTER YOU SHOULDERS LOOK UP MOUNTAIN YOU CAME FROM WHERE} \\
\textit{O} & So to gain that balance on your toe, all you have to do is center your shoulders by looking to the mountain. This is where you came from. \\
\textit{R} & To gain balance on your toe side, all you have to do is center your shoulders by looking up the mountain, back where you came from. \\
\midrule
\rowcolor{sechead}
\multicolumn{2}{@{}l}{\textbf{Example 2} --- BOBSL \textcolor{green!60!black}{\ding{51}}} \\
\textit{F} & \texttt{\small AN FORTUNATELY CATS BEST KEEP LITTLE PARROT ALIVE TAKE OUT} \\
\textit{O} & Unfortunately the cats are best kept outside to keep the little parrot alive. \\
\textit{R} & Unfortunately for the cats, the best way to keep this chubby little parrot alive is to take kitty out of the equation. \\
\midrule
\rowcolor{sechead}
\multicolumn{2}{@{}l}{\textbf{Example 3} --- How2Sign \textcolor{red!70!black}{\ding{55}}} \\
\textit{F} & \texttt{\small TODAY WE WORK LOWER HOUSE} \\
\textit{O} & Today we're working on the lower part of the house, the basement. \\
\textit{R} & Today we're going to work on stretching and strengthening the lower body. \\
\bottomrule
\end{tabular}\\[2pt]
{\scriptsize F = Fused pseudo-glosses, O = Our translation, R = Reference.}
\end{table}

\section{Ablation Studies}\label{sec:ablation}

All ablations use the full SignBind-LLM configuration with Flan-T5-XL on the How2Sign test set.

\begin{table*}[!t]
\centering
\scriptsize
\setlength{\tabcolsep}{4pt}
\renewcommand{\arraystretch}{1.06}

\begin{minipage}[t]{0.47\linewidth}
\centering
\captionof{table}{\textbf{Component ablation.}
Fusion ($-$13.5 B-4) and LLM ($-$14.3 B-4) cause the largest drops. Among inputs, lipreading is most important ($-$8.5 B-4). L-Acc = letter accuracy (\%).}
\label{tab:ablation}
\vspace{-2pt}
\begin{tabular}{@{}l cc@{}}
\toprule
\textbf{Variant} & \textbf{L-Acc} & \textbf{B-4} \\
\midrule
W/o Lipreading     & 54.7 & 14.6 \\
W/o Fingerspelling & 34.1 & 19.9 \\
W/o Fusion         & 50.2 &  9.6 \\
W/o Router         & 51.6 & 15.3 \\
W/o LLM            & 59.2 &  8.8 \\
\midrule
\rowcolor{bestblue}
\textbf{Full Model} & \textbf{73.6} & \textbf{23.1} \\
\bottomrule
\end{tabular}
\end{minipage}
\hfill
\begin{minipage}[t]{0.47\linewidth}
\captionof{table}{\textbf{Temporal alignment} between manual and non-manual streams. L-Acc = letter accuracy (\%).
Learned alignment outperforms all fixed offsets. Negative shifts help slightly (mouthings precede signs).}
\label{tab:alignment}
\vspace{-2pt}
\begin{tabular}{@{}l cc@{}}
\toprule
\textbf{Variant} & \textbf{L-Acc} & \textbf{B-4} \\
\midrule
No shift              & 68.7 & 18.2 \\
$\Delta{+}5$ frames   & 63.2 & 13.4 \\
$\Delta{-}5$ frames   & 71.0 & 19.1 \\
$\Delta{+}10$ frames  & 56.1 &  8.8 \\
$\Delta{-}10$ frames  & 65.3 & 16.6 \\
\midrule
\rowcolor{bestblue}
\textbf{Learned} & \textbf{73.6} & \textbf{23.1} \\
\bottomrule
\end{tabular}
\end{minipage}
\end{table*}

\noindent\textbf{Component ablation.}
Table~\ref{tab:ablation} reports performance when each component is individually removed. The fusion transformer and LLM produce the largest drops ($-$13.5 and $-$14.3 B-4), confirming that cross-modal reconciliation and linguistic reordering are both essential. Among inputs, lipreading has the strongest B-4 impact ($-$8.5), while removing fingerspelling causes the largest L-Acc.\ drop ($-$39.5\,pp) but a smaller B-4 decrease ($-$3.2), reflecting its localised role in proper noun recognition.

\noindent\textbf{Stand-alone expert evaluation.}
Table~\ref{tab:individual} isolates each stream. Continuous signing alone achieves only 7.3 B-4, which is expected considering the outputs of this stream are glosses and not english sentences. The lipreading branch reaches 66.0\% phoneme accuracy, confirming its role as the primary disambiguating signal. That neither stream alone approaches full-model performance highlights the complementarity of the architecture.


\begin{table*}[!t]
\centering
\scriptsize
\setlength{\tabcolsep}{4pt}
\renewcommand{\arraystretch}{1.06}
\begin{minipage}[t]{0.47\linewidth}
\centering
\captionof{table}{\textbf{Fusion and pre-training ablation.}
Top: fusion strategies. Cross-attention shows additional benefit while without gating the performance reduces significantly (concat+MLP). Bottom: We report zero-shot performance on How2Sign under various training setups.~\cite{YT-ASL}. pt = pre-trained T5, np = from scratch.}
\label{tab:fusion_pretrain}
\vspace{-2pt}
\begin{tabular}{@{}l c@{}}
\toprule
\textbf{Variant} & \textbf{B-4} \\
\midrule
\rowcolor{sechead}
\multicolumn{2}{@{}l}{\textit{Fusion strategy}} \\
\quad Concat + MLP              & 12.4 \\
\quad Gated Fusion           & 22.3 \\
\rowcolor{bestblue}
\quad \textbf{Gated Fusion + Cross Attn} & \textbf{23.1} \\
\addlinespace[3pt]
\rowcolor{sechead}
\multicolumn{2}{@{}l}{\textit{Zero-shot (YT-ASL $\rightarrow$ H2S)}} \\
\quad YT-ASL (np)~\cite{YT-ASL}   &  1.4 \\
\quad YT-ASL (pt)~\cite{YT-ASL}   &  4.0 \\
\quad \textbf{Ours}               &  \textbf{8.3} \\
\addlinespace[3pt]
\rowcolor{sechead}
\multicolumn{2}{@{}l}{\textit{How2Sign only (no YT-ASL)}} \\
\quad \'{A}lvarez et al.~\cite{How2Sign} & 2.2 \\
\quad GloFE-VN~\cite{lin-etal-2023-gloss} & 2.2 \\
\quad Tarr\'{e}s et al.~\cite{tarrés2023signlanguagetranslationinstructional} & 8.0 \\
\quad \textbf{Ours}               & \textbf{13.7} \\
\addlinespace[3pt]
\rowcolor{sechead}
\multicolumn{2}{@{}l}{\textit{Pre-train + Fine-tune}} \\
\quad YT-ASL (pt)~\cite{YT-ASL}   & 12.4 \\
\rowcolor{bestblue}
\quad \textbf{Ours}               & \textbf{23.1} \\
\bottomrule
\end{tabular}
\end{minipage}
\hfill
\begin{minipage}[t]{0.47\linewidth}
\centering
\captionof{table}{\textbf{Stand-alone expert encoder performance.}
Quality of CTC outputs from the individual expert streams for sign recognition and lipreading. Pho = phoneme accuracy (\%).}
\label{tab:individual}
\vspace{-2pt}
\begin{tabular}{@{}l cc@{}}
\toprule
\textbf{Stream} & \textbf{Pho} & \textbf{B-4} \\
\midrule
Sign only & --   & 7.3 \\
Lipreading only      & 66.0 & --  \\
\bottomrule
\end{tabular}

\vspace{6pt}
\includegraphics[width=\linewidth]{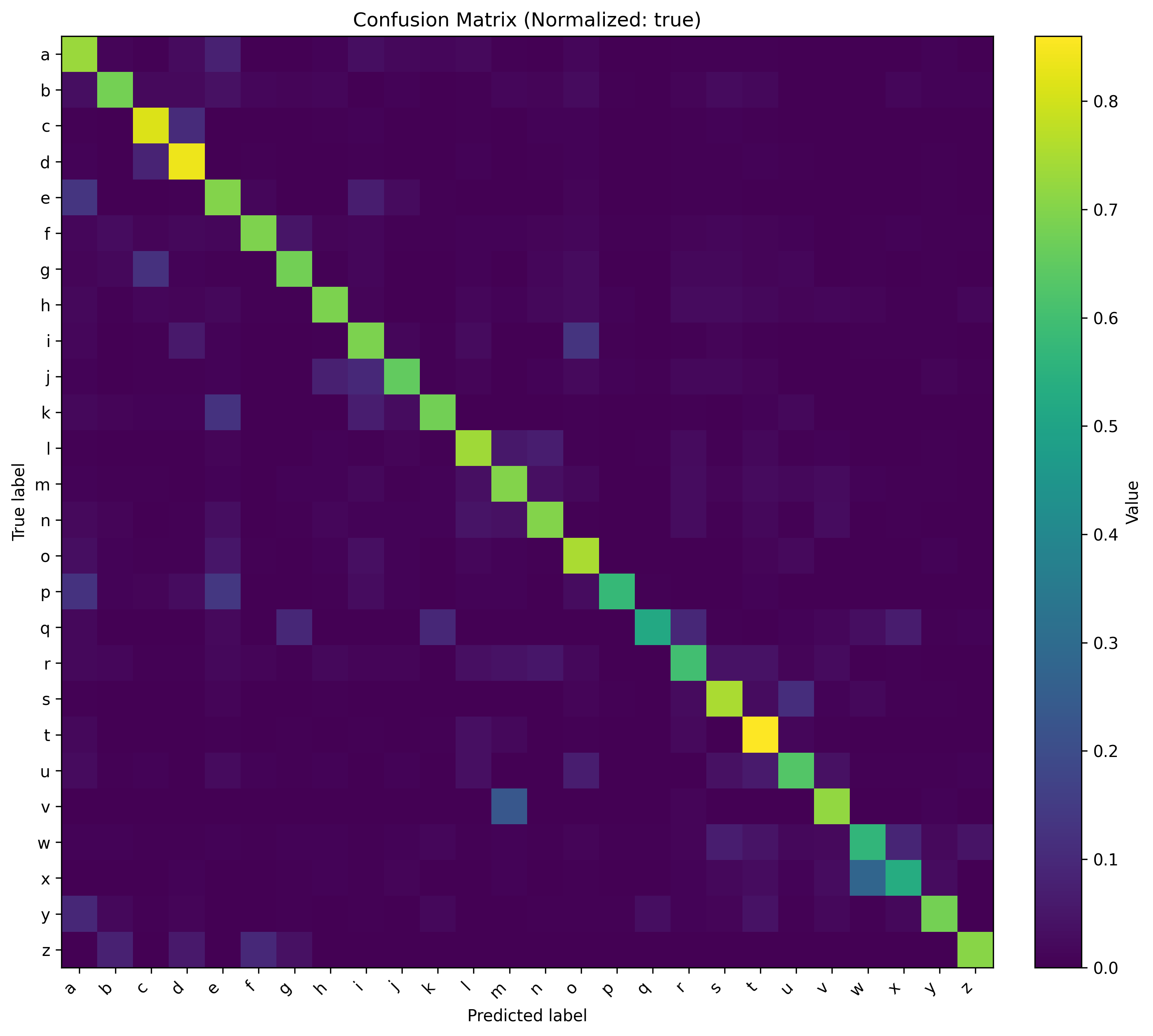}
\captionof{figure}{\textbf{Fingerspelling confusion matrix} on ChicagoFSWild+ (normalised)}
\label{fig:confusion}
\end{minipage}
\end{table*}

\noindent\textbf{Fusion strategy.}
Table~\ref{tab:fusion_pretrain} (top) compares three fusion strategies. Concatenation followed by a two-layer MLP achieves only 12.4 B-4---a 10.7-point drop---because it treats all time steps identically and cannot resolve the temporal misalignment between streams. Simply applying  gating and then performing concatenation before self-attention (without cross-attention between experts) achieves 23.1 B-4 (-1 B-4).

\noindent\textbf{Pre-training and zero-shot transfer.}
Table~\ref{tab:fusion_pretrain} (bottom three sections) compares training strategies against published baselines from~\cite{YT-ASL}. In the zero-shot setting (trained on YT-ASL, evaluated on How2Sign without fine-tuning), our model reaches 8.3 B-4---more than double the best YT-ASL baseline (4.0). Training only on How2Sign yields 13.7 B-4, which already exceeds all prior H2S-only methods including Tarr\'{e}s et al.\ (8.0). The full staged pipeline (pre-train on YT-ASL, fine-tune on H2S) achieves 23.1 B-4, nearly doubling the best comparable YT-ASL staged result (12.4). These results demonstrate that our expert architecture extracts more from the same data at every training configuration.

\noindent\textbf{Temporal alignment.}
Table~\ref{tab:alignment} compares fixed temporal offsets against learned alignment. A small negative shift ($\Delta{-}5$) slightly outperforms na\"ive frame-aligned fusion, consistent with mouthings preceding manual signs. Large positive shifts degrade performance severely ($\Delta{+}10$: $-$14.3 B-4), while negative shifts are more benign ($\Delta{-}10$: $-$6.5 B-4), suggesting greater robustness to mouthings leading hands than the reverse. The learned alignment mechanism outperforms all fixed offsets by a wide margin, validating the design choice of letting cross-attention learn the correspondence rather than assuming a fixed temporal relationship.

\noindent\textbf{Fingerspelling error analysis.}
Figure~\ref{fig:confusion} presents the confusion matrix for fingerspelling on ChicagoFSWild+. Interestingly, the dominant error patterns do not always involve visually similar handshapes: \texttt{M}/\texttt{V} (differing in finger extension). This suggests that performance may be limited by occlusion, resolution, or motion blur rather than the distance between fingers in the image, or that the fingerspelling network is leveraging additional context from the whole body to distinguish between letters. A detailed parts-of-speech analysis demonstrating that our improvements are concentrated on visually grounded lexical categories is provided in the supplementary material. 

\section{Conclusion}\label{sec:conclusion}

We have presented SignBind-LLM, a modular framework that decomposes sign language translation into dedicated experts for continuous signing, fingerspelling, and lipreading, fused with learned temporal alignment and decoded by a pre-trained LLM. This decomposition achieves state-of-the-art results on three benchmarks (\textbf{23.1} B-4 on How2Sign, \textbf{7.0} B-4 on BOBSL, \textbf{73.6\%} letter accuracy on ChicagoFSWild+) while requiring only 259 GPU hours, significantly less than existing approaches. Critically, the multi-stream architecture already outperforms prior methods at equal LM scale, confirming the gains are architectural. Ablations reveal complementary contributions from each stream, essential learned temporal alignment, and the advantage of modular training over end-to-end approaches. 

\section*{Acknowledgements}
This work was supported by EPSRC grant APP24554 (SignGPT-EP/Z535370/1), EPSRC grant APP78083 (UMCS UKRI3927) and through funding from Google.org via the AI for Global Goals scheme. The authors acknowledge the use of Isambard-AI National AI Research Resource (AIRR) funded by UK DSIT via UKRI and STFC [ST/AIRR/I-A-I/1023]. This work reflects only the authors' views and the funders are not responsible for any use that may be made of the information it contains.
We thank Oline Ranum for assistance with the parts-of-speech analysis.

\bibliographystyle{splncs04}
\bibliography{main}

\end{document}